\documentclass[a4paper,fleqn]{cas-sc}
\usepackage{array}
\usepackage{listings}
\usepackage[authoryear]{natbib}
\usepackage{float}
\usepackage{pifont}
\usepackage{multirow}
\def\tsc#1{\csdef{#1}{\textsc{\lowercase{#1}}\xspace}}
\tsc{WGM}
\tsc{QE}
\tsc{EP}
\tsc{PMS}
\tsc{BEC}
\tsc{DE}
\begin{document}
\let\WriteBookmarks\relax
\def\floatpagepagefraction{1}
\def\textpagefraction{.001}

% % Short title
% \shorttitle{Leveraging social media news}
\definecolor{codegreen}{rgb}{0,0.6,0}
\definecolor{codegray}{rgb}{0.5,0.5,0.5}
\definecolor{codepurple}{rgb}{0.58,0,0.82}
\definecolor{backcolour}{rgb}{0.95,0.95,0.92}

\lstdefinestyle{pythonstyle}{
    backgroundcolor=\color{backcolour},   
    commentstyle=\color{codegreen},
    keywordstyle=\color{blue}\bfseries,
    numbers=none,
    % numberstyle=\tiny\color{codegray},
    stringstyle=\color{codepurple},
    basicstyle=\ttfamily\footnotesize,
    breakatwhitespace=false,         
    breaklines=true,                 
    captionpos=b,                    
    keepspaces=true,                 
    % numbers=left,   % <-- remove or comment this
    numbersep=5pt,                  
    showspaces=false,                
    showstringspaces=false,
    showtabs=false,                  
    tabsize=4
}

% Apply the style
\lstset{style=pythonstyle}

% % Main title of the paper
\title [mode = title]{NormEval: A Unified Multi-Metric Framework for Evaluating Semantic Fidelity in Text Normalization}

\author[1]{Md Abdullah Al Kafi}[orcid=0009-0006-4361-029X]

\author[1]{Raka Moni}[orcid=0009-0008-5391-9893]
\author[2]{Walayat Hussain}[orcid=0000-0003-0610-4006]\cormark[1]
\ead{Walayat.Hussain@acu.edu.au}

\affiliation[1]{
  organization={Multidisciplinary Action Research (MARS) lab, Daffodil International University},
  addressline={Savar},
  city={Dhaka},
  postcode={1216},
  country={Bangladesh}
}

\affiliation[2]{
  organization={Artificial Intelligence for Decision Excellence (AIDX) Lab, Peter Faber Business School, Australian Catholic University},
  addressline={40 Edward Street},
  city={North Sydney},
  postcode={2060},
  country={Australia}
}

\begin{abstract}
Text normalization methods such as stemming and lemmatization are fundamental components of NLP pipelines. As new normalization tools are developed for diverse languages, evaluation methodologies remain fragmented, relying on Compression Ratio, downstream accuracy, or sequence-to-sequence prediction scores in isolation, failing to distinguish between beneficial vocabulary reduction and harmful semantic distortion. Moreover, text normalization underpins intelligent systems in high-stakes domains, including clinical decision support and legal document analysis, and principled evaluation methodology is essential. This paper proposes NormEval, a unified, multilingual evaluation framework comprising five complementary metrics: Compression Ratio (CR), Model Performance Delta (MPD), Information Retention Score (IRS), Algorithm Effectiveness Score (AES), and Average Normalized Levenshtein Distance (ANLD). These metrics assess normalization quality across three dimensions: macro-level efficiency, downstream utility, and micro-level morphological fidelity. The framework operationalizes a Safety Gate hypothesis: ANLD functions as an intrinsic structural hygiene check, utilizing character-level divergence ($\Delta$) to reveal aggressive mutations that macro-level embeddings and downstream tasks mask. Comprehensive ablation experiments on both Bangla and English datasets show that all the components are indispensable, and that the removal of any individual metric leads to a decrease in at least one evaluation aspect, which ultimately results in misleading algorithm rankings. Cross-lingual experiments on the XNLI benchmark in six typologically diverse languages show that our metrics are consistent and easy to interpret even under zero-shot transfer, and overall, these results demonstrate that NormEval is a standardised, multi-dimensional evaluation framework for text normalization that applies reliably across languages, tasks, and normalization methods.
\end{abstract}

% % % Research highlights
% \begin{highlights}
% \item This study proposes ``NormEval'', a unified multilingual evaluation framework for text normalization algorithms.

% \item The proposed ``Safety Gate'' mechanism detects the over-aggressive compression by monitoring the amount of distortion and anomalies at the word level.

% \item This study provides five complementary metrics, including the standard compression ratio, for the joint evaluation of macro-level efficiency, downstream utility, and semantic fidelity.
% \end{highlights}
% Keywords

\maketitle

\section{Introduction}

Natural language processing (NLP) has become a foundational technology across a rapidly growing range of intelligent systems, from clinical decision support and legal document analysis to sentiment-aware recommendation engines and multilingual search platforms \cite{Kowsher2022, Chauhan2023, nguyen_developing_2025, peral-garcia_comparing_2024, abulibdeh_natural_2025}.However, underlying every such pipeline lies a critical, yet underappreciated, preprocessing stage: \textit{text normalization}. 
Stemming, lemmatization, synonym merging, and other text normalization techniques reduce the variation in raw text, shrinking a high-dimensional feature space, and enabling the downstream models to generalize across different morphological forms of the same concept \cite{Jabbar2023, Balakrishnan2014, lovins1968development}. These steps have considerable practical consequences: the size of the vocabulary affects the memory footprint of vector space models, the efficiency and stability of neural network training, and the accuracy of information retrieval systems \cite{Paice1990, Fox1989, harman1991effective}.

Despite being the core of most NLP pipelines, the existing work on text normalization suffers from a serious methodological gap that has not received systematic attention: \textit{there is no standardized, comprehensive framework for evaluating whether a normalization algorithm is actually working correctly}. Since the seminal rule-based work of Lovins \cite{lovins1968development} and the widely adopted Porter algorithm \cite{porter2001snowball}, the field has proposed dozens of normalization techniques across English \cite{Paice1990, Fox1989}, Arabic \cite{Alnaied2020, AOtair2013, Alkhudari2020}, Bangla \cite{Dasgupta2007, Sarkar2008, sadia2019n, mugdha2024accurate}, Indonesian \cite{rianto2021improving, abidin2024text}, and other morphologically diverse languages \cite{Gadri2022, Munkova2021, agarwal2014morphological}.

The dominant evaluation paradigm in the normalization literature reduces to one of three isolated approaches, each of which is insufficient on its own.
The first and most prevalent is \textit{intrinsic compression measurement}: reporting the reduction in vocabulary size or feature dimensionality after normalization \cite{Adamson1974, Fox1989, Kasthuri2014, Islam2022}. The compression ratio indicates how much we save by using normalization, but it does not tell us whether the stems or lemmas are meaningful forms of words. An algorithm that truncates every word to its first three characters would score perfectly on compression while destroying all meaningful lexical content. The second method is \textit{extrinsic downstream evaluation}, where researchers evaluate the performance of classifiers or downstream tasks on the normalized text. If the classifier performs better, researchers conclude that the normalization was helpful, because the improved performance is a direct result of the normalization process, and it can be determined as the effectiveness of the normalization technique \cite{Al-Kabi2015, Dolamic2008, Saunack2021, Afrin2023, Islam2025}.
This approach is equally problematic. Downstream tasks are often a coarse, aggregate signal that conflates the effects of normalization with those of the model architecture, dataset characteristics, and class distribution. For example, a classifier may achieve higher accuracy on over-stemmed text simply because the aggressive reduction happens to align with the decision boundary of that particular model on that particular corpus, not because the normalization preserved linguistically meaningful distinctions \cite{Majumder2007, Paik2011}. Additionally, there is a trend of neural models to treat normalization as a \textit{sequence-to-sequence prediction problem} and evaluate it with token-level metrics such as BLEU or character accuracy \cite{Kowsher2022}.  It considers normalization evaluation as a generation task, and it requires high-quality gold standard lemmas, stems, or dictionary. These are scarce or nonexistent for most low-resource, morphologically rich languages because it also fails to reflect how the forms impact meaning or downstream tasks. However, the risks of such a strategy are not limited to the theoretical. Relying on just one score to evaluate a text normalization system can be risky. A system may appear accurate overall while quietly damaging the meaning of words.

This hidden problem is called semantic degradation: the gradual loss or distortion of meaning during processing \cite{Balakrishnan2014, sengupta1996morphological, agarwal2014morphological}.It is even more problematic with some languages that have rich morphology, i.e., there are plenty of suffixes/prefixes, and one word can be morphed into dozens of different forms, and stemming might merge two different words into a single word, which will negatively impact precision/recall. In Bangla, a highly inflected and widely spoken language, this is particularly important. Existing tools such as BNLTK \cite{bnltk} often remove suffixes aggressively, achieving good compression but producing invalid stems. These issues are not reliably detected by compression or classification accuracy alone \cite{mugdha2024accurate, RaniBiswas2024}.  Analogous problems have been documented for Arabic \cite{Alnaied2020}, Indonesian \cite{abidin2024text}, and South Asian languages more broadly \cite{Gadri2022}, suggesting a systemic failure in the evaluation methodology currently used rather than a language or algorithm-specific problem.

Another challenge is that even studies that report multiple metrics rarely consider the relationship between them. Compression and accuracy are frequently reported in the same paper, but rarely analyzed as potentially conflicting signals: a result that looks like a success on one axis may in fact indicate a failure on another. For example, Majumder et al. \cite{Majumder2007} provide a partial trade-off analysis, and even that work does not incorporate semantic preservation or micro-level morphological distortion as explicit evaluation dimensions. As our analysis of the literature in Table~\ref{tab:comparison_part1_sorted} shows, across all reviewed normalization studies, not a single prior work simultaneously evaluates intrinsic compression, downstream utility, semantic fidelity, micro-level morphological distortion, and statistical robustness. This gap makes it difficult to compare normalizers, and we do not see where they fail. It is hard to determine whether a normalizer improves or harms an NLP system.

To address this critical gap, this paper proposes \textbf{NormEval}, a unified, data-driven evaluation framework for text normalization that integrates five complementary metrics into a single, coherent assessment pipeline.
NormEval operationalizes three orthogonal dimensions of normalization quality: (1) \textit{macro-level efficiency}, quantified by the Compression Ratio (CR) and the Algorithm Effectiveness Score (AES), which combines CR with semantic preservation into a single harmonic measure; (2) \textit{downstream utility}, captured by the Model Performance Delta (MPD) with mandatory statistical significance testing; and (3) \textit{semantic fidelity}, assessed at the macro level by the Information Retention Score (IRS), computed via contextual transformer embeddings, and at the micro level by the Average Normalized Levenshtein Distance (ANLD), a character-level diagnostic that functions as a \textit{safety gate} against destructive over-stemming. The framework was tested on several languages and normalization algorithms, and was shown to generalize well and perform consistently across different language families, domains, and data distributions.

The primary contributions of this work are as follows:
\begin{enumerate}
    \item \textbf{A unified five-component evaluation framework (NormEval)} that includes efficiency, downstream performance, semantic preservation, and morphological integrity within a single framework.
    \item \textbf{The Safety Gate formalism}, using Average Normalized Levenshtein Distance (ANLD) thresholds, which helps detect harmful over-stemming that may not appear in embedding metrics or accuracy scores.

    \item \textbf{Empirical evidence that downstream model accuracy, or F1-score, is an unreliable proxy for normalization quality},  shown by non-significant performance differences (p $>$ 0.05) across models and languages, suggesting Model Performance Delta (MPD) is better used as a “hygiene” check rather than a performance measure.

    \item \textbf{The multidimensional cross-linguistic normalization evaluation} using the same standardized framework, the study shows that tools with identical Algorithm Effectiveness Scores (AES) can still differ in their morphological risk, since they are evaluated under the same test conditions.

    % \item \textbf{Publicly available code and data} for reproducing all experiments and adapting NormEval to new languages and normalization algorithms, available at \url{https://github.com/abkafi1234/stemming}. The tool can be used by simply "pip install normeval."

    \item \textbf{A fully reproducible open-source ecosystem} consisting of all evaluation datasets, benchmarks, and the core Python package interface. To preserve the integrity of the double-blind review process, the source code repositories and package installation links have been anonymized for this submission.
\end{enumerate}

The rest of the paper is organized as follows. Section~\ref{sec:related} reviews prior work on text normalization and its evaluation. Section~\ref{sec:methodology} defines the NormEval framework and the experimental setup. Section~\ref{sec:results} presents results on multiple datasets. Section~\ref{sec:conclusion} concludes with key findings and future directions.

\section{Background and Related Work}
\label{sec:related}

Text normalization is a key step in nearly all NLP pipelines, and its decisions affect the operation of all downstream components, but despite this importance, text normalization has not been systematically evaluated. In this section, we describe existing normalization methods and their typical evaluation, and the weaknesses of current practice that our framework addresses.

%%------------------------------------------------------------
\subsection{Text Normalization: Foundations and Algorithmic Landscape}
%%------------------------------------------------------------

The history of text normalization can be traced back to the late 1960s, when Lovins presented a stemming algorithm for the English language that is based on removing suffixes to bring words to a base form, and most stemming approaches over the next 50 years inherited this approach \cite{lovins1968development}. Over the next few decades, more sophisticated rule-based stemmers were developed, for example, Fox et al. introduced the notion of conflation classes to measure the degree to which a stemmer can group word variants\cite{Fox1989}. Paice et al. later expanded on Lovins' work with the Lovins–Paice stemmer, which iteratively applies a set of suffix stripping rules to more effectively balance aggressive conflation with linguistic accuracy \cite{Paice1990}. Adamson and Boreham introduced association measures to assess stemming quality, an early shift from rule design to empirical evaluation. However, their focus was still mainly on vocabulary reduction \cite{Adamson1974}.

In the 2000s, text normalization methods became more diverse, largely due to advances in information retrieval research and the availability of large text datasets. Kumar et al. reviewed the principal stemming algorithms and their performance characteristics, while Husain  conducted a comparative study of English stemmers, exposing inconsistencies in how the same algorithm performs across different corpora and tasks \cite{Kumar2010, Husain2012}.
A key finding from Balakrishnan et al. is that there is no single stemming method that works best for all situations. How much of a word's ending should be removed depends on the task, the text collection, and the language being used \cite{Balakrishnan2014}.
However, this insight has been reflected very rarely in practice, because most papers report a single performance metric on a single dataset with little to no discussion of how the results generalize across tasks or settings.

%%------------------------------------------------------------
\subsection{Stemming and Lemmatization Across Languages}
%%------------------------------------------------------------

Text normalization is much more complex for morphologically rich languages, where one lexical root can have many different forms, because poorly designed stemming can inappropriately collapse words with different meanings into the same token, thus both harming linguistic fidelity and downstream performance. Jabbar et al. provide a comprehensive longitudinal survey spanning stemming research from 1968 to 2023, documenting how the field has progressively expanded from English-centric methods to a broad multilingual landscape, while simultaneously observing that evaluation rigor has not kept pace with algorithmic diversity \cite{Jabbar2023}.

\textbf{Arabic and Semitic languages} have attracted substantial normalization research due to their highly complex root-and-pattern morphology. Alnaied et al. proposed AMIR, a multi-level Arabic normalization system integrating both stemming and lemmatization, building on earlier algorithmic contributions from Otair and Alkhudari \cite{Alkhudari2020, Alnaied2020, AOtair2013}.
Al-Kabi et al. evaluated Arabic stemmers using only sentiment classification accuracy, without assessing stemming quality or meaning preservation. This approach is common in multilingual normalization research \cite{Al-Kabi2015}.

\textbf{Bangla} language has rich morphological forms, and there are not many resources available for processing Bangla. An unsupervised Bangla stemmer was developed by Dasgupta and Ng, and a rule-based Bangla stemmer was developed by Sarkar.\cite{Dasgupta2007, Sarkar2008}.Majumder et al. conducted one of the most comprehensive Bangla stemming studies by evaluating both compression and retrieval performance. However, they did not assess semantic meaning preservation or detailed morphological errors \cite{Majumder2007}.
More recently, Hoque et al. and Chakrabarty et al. improved Bangla stemming algorithms, and they developed these algorithms for information retrieval and text categorization tasks \cite{Hoque2016, Chakrabarty2016}. Islam et al. proposed a rule-based approach for morphological analysis of Bangla words \cite{ Islam2022}. Sadia et al. proposed an n-gram–based approach to Bangla stemming, and Mugdha et al. specifically designed a stemmer for text classification tasks, recognizing that task-specific tuning may outperform general-purpose normalization, but without providing a principled framework for measuring the trade-off involved \cite{sadia2019n, mugdha2024accurate}. Sengupta et al. performed spelling correction for Bangla words, and Biswas et al. developed fundamental resources and a linguistic corpus for further research on Bangla NLP \cite{sengupta1996morphological, RaniBiswas2024}. For Bangla, Afrin et al. and Islam et al. study lemmatisation for text classification and sentiment analysis, respectively, and they only evaluate their lemmatisers using downstream accuracy. Neither of these studies reports intrinsic or semantic quality metrics for the generated lemmas because they focus on the aforementioned applications and do not provide a comprehensive analysis of their lemmatisers \cite{Afrin2023, Islam2025}.

\textbf{South and Southeast Asian languages} more broadly have received growing normalization attention.
Agarwal et al. \cite{agarwal2014morphological} proposed a rule-based morphological analyzer for Hindi, while Chauhan et al. \cite{Chauhan2023} developed a DFA-based lemmatizer for Gujarati topic modeling, evaluating the system through topic coherence scores rather than morphological accuracy.
For Indonesian, Rianto et al. examined stemming in informal social media text, Giri et al. proposed the MTstemming algorithm for multi-stage preprocessing pipelines, and Abidin et al. conducted a comprehensive systematic survey of stemming and lemmatization for local Indonesian languages, identifying the scarcity of standardized evaluation benchmarks as a primary limitation of the field \cite{rianto2021improving, giri2021, abidin2024text}. Studies by Gadri and Moussaoui, and Munkova et al. on highly inflectional languages showed that as morphological complexity increases, thorough evaluation becomes less common \cite{Gadri2022, Munkova2021}.

\textbf{Neural and hybrid approaches} represent the most recent development in the normalization landscape. Kowsher et al. argued that Bangla-BERT can reduce the need for stemming because transformer models can learn different word forms through their attention mechanisms \cite{Kowsher2022}. Tan et al. combined RoBERTa with LSTM architectures for sentiment analysis and demonstrated competitive performance without explicit normalization \cite{9716923}.
These findings raise an important issue: since present deep learning models are already capable of absorbing some degree of surface variation, if one only uses downstream accuracy as a metric for a normalizer, it is difficult to disentangle its contribution. Pal et al. and Qureshi et al. offer a more detailed discussion of semantic processing and multilingual normalization, arguing that the evaluation protocols should take into account the distinction between normalization quality and model capacity \cite{RanjanPal2015, Qureshi2021}.

%%------------------------------------------------------------
\subsection{Evaluation Methodologies: Three Isolated Paradigms}
%%------------------------------------------------------------

The survey reveals that most of the current normalization methods have only been evaluated with three methods, each measuring one dimension of quality and ignoring others. In this section, we will review these methods in detail.

\textbf{Paradigm 1: Intrinsic compression metrics.}
The oldest and most widely used evaluation approach measures how much a normalizer reduces the vocabulary of a corpus, typically reported as vocabulary compression ratio, type-token ratio reduction, or the size of the resulting conflation classes \cite{Adamson1974, Fox1989, Kasthuri2014, Islam2022, sadia2019n}. The intuition is straightforward: if the goal of normalization is to reduce lexical variability, then the amount of vocabulary reduction is a direct indicator of its performance. However, this approach is flawed because it rewards excessive simplification without checking meaning. A system may score high by over-compressing the vocabulary to the point of collapsing semantically unrelated words to the same form, resulting in compression metrics that cannot distinguish between correct and incorrect merging of word forms.\cite{Balakrishnan2014, Paice1990}.

\textbf{Paradigm 2: Extrinsic downstream accuracy.}
A second widely adopted paradigm evaluates normalization indirectly, by measuring the accuracy of a classification or retrieval model trained on normalized text and comparing it to a baseline trained on unnormalized text \cite{Al-Kabi2015, Dolamic2008, Hoque2016, Chakrabarty2016, Saunack2021, Afrin2023, Islam2025}. This approach is appealing because it measures practical usefulness, but it also introduces several important methodological problems.
First, downstream accuracy is a coarse aggregate signal that conflates the normalizer's contribution with the model architecture, hyperparameter choices, class distribution, and dataset characteristics. A bad normalizer may look good if the classifier is strong enough to ``overpower'' the errors of the normalizer, while a good normalizer may look bad if paired with a badly tuned classifier \cite{Paik2011}.
Second, and critically, downstream accuracy is not a diagnostic procedure. A small gain in accuracy does not show which linguistic changes caused the improvement, and it gives no insight into performance on rare classes, ambiguous forms, or unseen inputs.

Third, without statistical significance testing, small performance gains are often treated as meaningful improvements. However, as shown in Section~\ref{sec:results}, many of these gains do not hold up under proper hypothesis testing \cite{Majumder2007, Paik2011}.

\textbf{Paradigm 3: Sequence prediction and generation scores.}
The most recent line of work, inspired by neural NLP methods, treats normalization as a sequence-to-sequence generation task, and predicted forms are compared to reference outputs using metrics such as token accuracy, BLEU, and character-level edit distance \cite{Kowsher2022}. While this method is more linguistically informed than compression-based evaluation, it depends on high-quality annotated datasets. Such resources are often unavailable for many languages, including several in South and Southeast Asia.\cite{RaniBiswas2024, abidin2024text}. Additionally, these metrics compare the output of models to a single fixed reference and only evaluate compliance to a single selected normalization standard, but they say little about downstream usefulness or semantic fidelity, because a stem that differs from the reference might still be semantically identical, and a reference-perfect stem may still be overly conflating different words in a particular domain.

%%------------------------------------------------------------
\subsection{The Semantic Fidelity Blind Spot}
%%------------------------------------------------------------

Across all three evaluation approaches, a key aspect is missing: direct measurement of whether normalization preserves meaning. This is a significant gap. Semantic preservation is essential to normalization, since grouping word forms is only valid if those forms are truly related in meaning. When a stemmer produces over-truncated, linguistically invalid base forms, as BNLTK does for Bangla \cite{mugdha2024accurate} or as aggressive English stemmers do for technical vocabulary \cite{Paice1990}, the resulting representation is not a semantically compressed version of the original text; it is a semantically corrupted one. Vocabulary reduction and downstream accuracy are not sufficient indicators of this problem because a bad normalization method may still reduce the vocabulary or even improve task performance by removing superficial noise instead of meaningful content. Edit distance captures surface differences but depends on reference forms, not meaning, and therefore does not show the real impact on downstream performance.

The need for evaluating semantic fidelity becomes more compelling in morphologically complex and low-resource languages. For Bangla, where a root word may have 50 or more grammatically valid inflected forms \cite{sengupta1996morphological, RaniBiswas2024}, an incorrect stem selection may combine two unrelated words, eliminating their distinction, and it also results in the loss of the meaning of both words. Normalization should be verified to be safe on two levels: Globally, with respect to the similarity between the spaces, and locally, with respect to the change induced per word. As Balakrishnan et al. noted, the lack of semantic evaluation has limited the field’s ability to determine when normalization helps or harms performance. This makes it difficult to make reliable, evidence-based choices in practical NLP systems \cite{Balakrishnan2014}.

%%------------------------------------------------------------
\subsection{Research Gap and Positioning of This Work}
%%------------------------------------------------------------

The analysis above reveals a structural gap in the normalization evaluation literature that is both deep and remarkably consistent across languages, tasks, and time periods.
As Table~\ref{tab:comparison_part1_sorted} summarizes, across all reviewed works, spanning more than five decades from Lovins to Islam et al., no prior work has investigated intrinsic compression efficiency, downstream task utility, macro-level semantic fidelity, micro-level morphological distortion, and statistical robustness in a unified framework \cite{lovins1968development, Islam2025}. The closest approximation is Majumder et al., who combine compression and retrieval metrics with partial statistical testing, but without any semantic fidelity measurement \cite{Majumder2007}. Most studies evaluate normalization using only one or two of the five key dimensions. This causes three problems: First, comparisons between algorithms are not meaningful because of the lack of consistent metrics, and second, we cannot accurately diagnose mistakes. We cannot identify the cause of performance degradation, and third, poor normalization schemes remain in circulation because existing tests fail to demonstrate the adverse effects.

This work fills this gap by presenting NormEval, a unified five-component evaluation framework that captures all five dimensions within a single reproducible assessment pipeline. By integrating Compression Ratio (CR), Information Retention Score (IRS), Algorithm Effectiveness Score (AES), Model Performance Delta (MPD) with mandatory significance testing, and Average Normalized Levenshtein Distance (ANLD) as a morphological safety gate, NormEval provides a complete and internally consistent basis for evaluating whether a normalization algorithm is genuinely improving an NLP system or covertly degrading it.

\begin{table}[ht]
\centering
\small
\setlength{\tabcolsep}{4pt}
\caption{Comparison of stemming and Lemmatization Method Properties}
\label{tab:comparison_part1_sorted}
\begin{tabular}{p{3.0cm}ccccc}
\toprule
\textbf{Study} & \textbf{Intrinsic Metric} & \textbf{Extrinsic Evaluation} & \textbf{Semantic Fidelity} & \textbf{Statistical Validation} & \textbf{Trade off} \\
\midrule
\cite{lovins1968development} & $\times$ & $\times$ & $\times$ & $\times$ & $\times$\\
\cite{Adamson1974} & $\checkmark$ & $\times$ & $\times$ & $\times$  & $\times$  \\
\cite{Paice1990} & $\checkmark$ & $\checkmark$ & $\times$ & $\times$ & $\times$ \\
\cite{Fox1989} & $\checkmark$ & $\times$ & $\times$ & $\times$ & $\times$ \\
\cite{Al-Kabi2015} & $\times$ & $\checkmark$ & $\times$ & $\times$ & $\times$ \\
\cite{Dasgupta2007} & $\checkmark$ & $\checkmark$ & $\times$ & $\times$ & $\times$ \\
\cite{Majumder2007} & $\checkmark$ & $\checkmark$ & $\times$ & $\checkmark$ & $\checkmark$ \\
\cite{Sarkar2008} & $\checkmark$ & $\checkmark$ & $\times$ & $\times$ & $\times$ \\
\cite{Dolamic2008} & $\times$ & $\checkmark$ & $\times$ & $\times$ & $\times$ \\
\cite{Kumar2010} & $\times$ & $\times$ & $\times$ & $\times$  & $\times$  \\
\cite{Qureshi2021} & $\checkmark$ & $\checkmark$ & $\times$ & $\times$ & $\times$\\
\cite{Paik2011} & $\checkmark$ & $\checkmark$ & $\times$ &  $\checkmark$ & $\times$ \\
\cite{Kasthuri2014} & $\checkmark$ & $\times$ & $\times$ & $\times$  & $\times$ \\
\cite{Husain2012} & $\checkmark$ & $\checkmark$ & $\times$ & $\times$  & $\times$  \\
\cite{Hoque2016} & $\checkmark$ & $\checkmark$ & $\times$ & $\times$  & $\times$  \\
\cite{sadia2019n} & $\checkmark$ & $\times$ & $\times$ & $\times$ & $\times$ \\
\cite{Chakrabarty2016} & $\checkmark$ & $\checkmark$ & $\times$ & $\times$  & $\times$ \\
\cite{Saunack2021} &  $\times$ & $\checkmark$ & $\times$ & $\times$  & $\times$\\
\cite{Islam2022} & $\checkmark$ & $\times$ & $\times$ & $\times$  & $\times$\\
\cite{Afrin2023} & $\times$ & $\checkmark$ & $\times$ & $\times$  & $\times$ \\
\cite{Islam2025} & $\times$ & $\checkmark$ & $\times$ & $\times$  & $\times$  \\
\midrule
\textbf{Our Method} & $\checkmark$ & $\checkmark$ & $\checkmark$ & $\checkmark$ & $\checkmark$ \\
\bottomrule
\end{tabular}
\end{table}

\section{Methodology}
\label{sec:methodology}
Our evaluation framework assesses text normalization along three complementary dimensions: macro-level efficiency, downstream utility, and micro-level morphological fidelity of normalized inputs. We evaluated our approach in three settings: (i) a clean parallel corpus for an ablation study, (ii) a sentiment review dataset to verify the morphological challenges, and (iii) a multilingual test set XNLI, used to evaluate cross-lingual performance.

\subsection{Dataset Preparation}

\subsubsection{Dataset 1: Bangla Text Sentence Dataset (Complexity Corpus)}
The primary dataset is a sentence-level parallel corpus with syntactic labels (Simple, Compound, Complex), taken from the Bangla Text Sentence Dataset (BTSD) \cite{Das2023}. In the scope of our work, we focus on the Bangla source and its English translations, which provide a direct and controlled way to benchmark structural classification across the two languages. The classes are well balanced (Simple: 888, Compound: 892, Complex: 879); therefore, the macro F1-score is a suitable metric to measure the performance, and the corpus is parallel. This isolates the impact of morphological variation. This is valuable for investigating stemming and lemmatization. This corpus serves as our primary evaluation platform for the NormEval ablation analysis, as it provides a suitable basis for comparison.

\subsubsection{Dataset 2: Bangla Sentiment Classification Corpus}
In order to further evaluate the robustness of the proposed framework in a more realistic and challenging setting, we also make use of an auxiliary sentiment classification dataset \cite{Biswas2025} with 9,163 examples (Positive: 4,139; Negative: 2,826; Neutral: 2,189), published in Mendeley.Unlike the complexity corpus, this dataset is intentionally imbalanced and contains a range of text types (news, social media, blogs), because we retain the natural class imbalance of this dataset to test the normalization techniques under realistic conditions and examine their impact on the performance of the minority class. We use it in our experiments as a robustness testbed to ensure that the framework's results generalize outside of controlled settings.

\subsubsection{Dataset 3: Cross-Lingual Natural Language Inference (XNLI)}
To evaluate the cross-lingual generalizability of the NormEval framework, we incorporate the Cross-Lingual Natural Language Inference (XNLI) benchmark \cite{conneau2018xnli}. XNLI is a widely adopted multilingual evaluation dataset that extends the MultiNLI corpus \cite{williams2018broad} to 15 languages by providing human-translated premise–hypothesis pairs with three inference labels: \textit{entailment}, \textit{neutral}, and \textit{contradiction}. For this study, we select six typologically diverse languages, such as English, French, Spanish, German, Arabic, and Russian, to represent a broad spectrum of morphological complexity, script diversity, and linguistic typology. For each language, we use the official XNLI validation split (2,490 examples per language) as the evaluation corpus, and the English MultiNLI training split (392,702 examples) to train the downstream classifier. The XNLI corpus serves exclusively as the multilingual generalizability testbed, allowing us to observe how normalization behavior and metric profiles vary across languages with fundamentally different morphological structures.

\begin{figure*}[ht]
    \centering
    \includegraphics[width=0.8\textwidth]{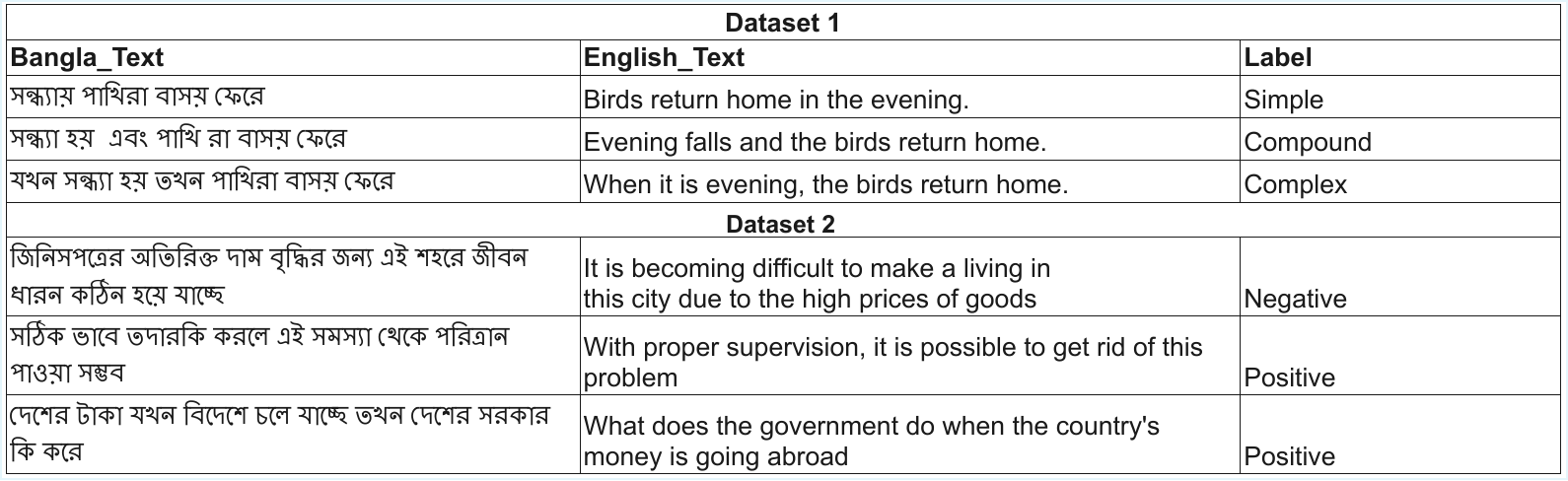}
    \caption{Dataset Examples}
    \label{fig:DatasetExample}
\end{figure*}

An illustrative example of the first two dataset formats used in this study is provided in Figure~\ref{fig:DatasetExample}.

\subsection{Experimental Design}
To ensure that each dataset is evaluated in a manner consistent with its intended role, we define three distinct experimental protocols, each targeting a specific evaluation objective of the NormEval framework. Additionally, we used the Wilcoxon signed-rank test to assess whether observed differences between paired evaluation outcomes were statistically significant without assuming normality of the underlying score distributions, making it appropriate for comparing model performance across repeated measurements and heterogeneous dataset conditions \cite{wilcoxon_individual_1992}. Table~\ref{tab:exp_design} provides a summary of the mapping between datasets, normalization methods, metrics evaluated, and experimental goals. 

\begin{table}[ht]
\centering
\caption{Summary of Experimental Design}
\label{tab:exp_design}
\small
\setlength{\tabcolsep}{4pt}
\begin{tabular}{p{2.2cm} p{2.5cm} p{2.5cm} p{3.5cm}}
\toprule
\textbf{Dataset} & \textbf{Normalization Methods} & \textbf{Metrics Evaluated} & \textbf{Experimental Goal} \\
\midrule
XNLI (6 languages) & Unicode NFKC + Snowball (language-specific) & CR, IRS, AES, ANLD, MPD & Cross-lingual generalizability evaluation \\
\midrule
BTSD Complexity (EN + BN) & Porter, Snowball, WordNet, SpaCy (EN); BNLTK, BaNeL, BanLemma, BanglaLem (BN) & CR, MPD, IRS, AES, ANLD & Ablation study: necessity of each metric \\
\midrule
Sentiment Analysis (BN) & Snowball, BNLTK, BanLemma & CR, IRS, AES, ANLD, MPD & Robustness under domain shift and class imbalance \\
\bottomrule
\end{tabular}
\end{table}

\subsubsection{Experiment 1: Multilingual Generalizability Evaluation (XNLI)}

For the multilingual evaluation, the XNLI benchmark \cite{conneau2018xnli} is used. The evaluation is restricted to the six languages for which a Snowball stemmer is available: Arabic (ar), German (de), English (en), Spanish (es), French (fr), and Russian (ru). Languages without a Snowball stemmer were excluded because, for such languages, the normalization pipeline reduces to Unicode NFKC normalization, lowercasing, diacritic stripping, and punctuation removal only, operations that do not produce meaningful morphological compression and therefore yield CR and ANLD values that reflect tokenization artifacts rather than genuine normalization behavior. For each language, the official XNLI validation split (2,490 examples per language) is used as the evaluation corpus, and the English MultiNLI training split (392,702 examples) is used to train the downstream classifier. This setup reflects a standard zero-shot cross-lingual transfer protocol: the classifier is trained once on English and evaluated on each target language without any language-specific fine-tuning, with cross-lingual generalization enabled entirely through the multilingual sentence encoder.

A standardized normalization pipeline is applied uniformly and independently to both the premise and hypothesis of each NLI pair before encoding. The pipeline consists of five sequential steps: (1) Unicode NFKC normalization for script standardization, (2) lowercasing with Turkish-aware dotless-i handling, (3) diacritic stripping for Latin-script languages, (4) punctuation removal including extended Arabic punctuation, and (5) language-specific Snowball stemming.

Since XNLI is a relational inference task, encoding the premise and hypothesis as a single concatenated string would lose the directional entailment signal. Instead, premises and hypotheses are encoded separately using the \texttt{paraphrase-multilingual-MiniLM-L12-v2} model \cite{reimers2019sentence}, which produces 384-dimensional sentence embeddings via mean pooling over token representations. A four-part InferSent-style interaction feature vector is then constructed as $[\mathbf{u},\, \mathbf{v},\, |\mathbf{u}-\mathbf{v}|,\, \mathbf{u} \odot \mathbf{v}]$, where $\mathbf{u}$ and $\mathbf{v}$ are the $\ell_2$-normalized premise and hypothesis embeddings respectively, yielding a 1,536-dimensional feature vector per example. A Logistic Regression classifier (multinomial softmax, $C=1.0$, \texttt{lbfgs} solver, \texttt{max\_iter}=1000) is trained once on the English MultiNLI training features and evaluated on each language's validation split without retraining.

The MPD for each language is computed as the difference in macro F1-score between the original and normalized validation sets: $\text{MPD} = \text{F1}_{\text{norm}} - \text{F1}_{\text{orig}}$. The remaining four NormEval metrics (CR, IRS, AES, ANLD) are computed on the concatenated premise hypothesis texts per language, as these metrics operate at the corpus level and do not require the relational NLI structure. Results are reported in Table~\ref{tab:multilingual_nli_results} in Section~\ref{sec:results}.

\subsubsection{Experiment 2: Ablation Study (BTSD Complexity Corpus)}
The ablation study is conducted on the BTSD complexity corpus using both the English and Bangla partitions. Four standard machine learning classifiers, such as Logistic Regression, Multinomial Naive Bayes (MNB), Support Vector Machine (SVM), and Random Forest (RF), are used as downstream models, all with default hyperparameters as implemented in scikit-learn to ensure reproducibility and avoid bias from model-specific tuning. For each normalization method, all five NormEval metrics are computed. To isolate the contribution of each metric, we systematically evaluate what information would be lost if any single metric were removed from the framework, demonstrating that no metric is redundant and that each captures a distinct and non-substitutable dimension of normalization quality. Statistical significance of MPD values is assessed using the Wilcoxon signed-rank test ($\alpha = 0.05$) across classifier predictions on the original and normalized corpora.

\subsubsection{Experiment 3: Robustness Evaluation (BTSD Sentiment Corpus)}
The robustness evaluation is conducted on the Bangla sentiment corpus using three normalization methods: Snowball, BNLTK, and BanLemma \cite{Biswas2025, bnltk, sadia2019n, porter2001snowball}. The same four classifiers and default hyperparameter settings from Experiment 2 are retained to ensure comparability. This experiment specifically tests whether the ANLD-based safety gate hypothesis that morphological distortion beyond a critical threshold leads to measurable downstream performance degradation holds under domain shift and class imbalance. Macro F1-score is reported to account for the imbalanced class distribution.

\subsection{Proposed Framework}
This framework aims to evaluate normalization methods, specifically stemming and lemmatization, from three complementary perspectives: Macro-level Effectiveness, Impact on Downstream Tasks, and Semantic Fidelity. Using five evaluation metrics provides a holistic assessment that avoids the pitfalls of disconnected single-metric-based evaluation and reveals both the advantages and the hidden problems of each normalization strategy.

\subsubsection{Component 1: Compression Ratio}
The first component addresses the macro-level effects of normalization, quantifying them through vocabulary reduction. The percentage reduction in vocabulary is defined as:
\begin{equation}
    CR = \frac{\text{\# Unique Words Before Transformation}}{\text{\# Unique Words After Transformation}}
\end{equation}
Here, the symbol ``\#'' represents the total count of unique tokens. \begin{itemize}
    \item $CR > 1.0$: represents a more successful vocabulary compression; a larger reduction percentage indicates more aggressive stemming.
    \item $CR = 1.0$: indicates that there is no change in vocabulary size (ineffective compression).
    \item $CR < 1.0$: represents vocabulary expansion, a rare anomaly where the process increases the number of unique tokens (e.g., due to erratic tokenization).
\end{itemize}
Here, we kept the compression ratio unbounded to make the difference visible.

\subsubsection{Component 2: Model Performance Delta (MPD)}
The Model Performance Delta (MPD) quantifies the impact of normalization on downstream task performance, adapted from Harman \cite{harman1991effective}.  It is defined as the difference between the performance of a model trained and evaluated on normalized text versus the same model on the original text:

\begin{equation}
MPD = \mathcal{P}_{\text{norm}} - \mathcal{P}_{\text{orig}}
\end{equation}

where $\mathcal{P}$ denotes any task-appropriate performance measure accuracy, macro F1, BLEU, ROUGE, or others. A positive MPD indicates that normalization improves downstream performance; a negative MPD indicates degradation; and a value near zero indicates task-neutrality. MPD is intentionally defined as a generic slot: the practitioner substitutes the metric appropriate to their task (classification, sequence labeling, machine translation, summarization, etc.), making NormEval applicable beyond any single NLP paradigm.

In this study, MPD is instantiated as the difference in macro F1-score on a text classification task, serving as a case study to demonstrate the metric's diagnostic behavior. The generalization of MPD to other task types, including sequence labeling, question answering, and generative tasks, is discussed in Section~\ref{sec:limitations}.

\subsubsection{Component 3: Information Retention Score (IRS)}
The IRS measures how well the document's meaning is preserved after stemming by computing the semantic similarity between the original document and its stemmed/normalized version. We compute this as the cosine similarity between the two aggregated embedding vectors. Higher IRS means that the normalized text is closer in meaning to the original document.

\begin{equation}
    IRS = \cos(\theta) = 
    \frac{V_{\text{orig}} \cdot V_{\text{norm}}}
    {\|V_{\text{orig}}\| \,  \|V_{\text{norm}}\|}
\end{equation}

Here, $V_{\text{orig}}$ and $V_{\text{norm}}$are the sentence/document level vectors obtained by averaging the contextualized token embeddings, which were generated using transformer-based models, specifically \texttt{csebuetnlp/banglabert} for Bangla,  \texttt{distilbert-base-uncased} for English and \texttt{sentence-transformers/paraphrase-multilingual-MiniLM-L12-v2} for XNLI dataset.

\subsubsection{Component 4: Algorithm Effectiveness Score (AES)}

The Algorithm Effectiveness Score (AES) provides a unified measure of normalization quality by combining vocabulary compression and semantic preservation into a single bounded score. A well-performing normalization algorithm should simultaneously reduce vocabulary size (efficiency) and retain the semantic content of the original text (fidelity). Optimizing for only one dimension is insufficient: an algorithm that compresses aggressively but destroys meaning is harmful, and an algorithm that preserves meaning perfectly but changes nothing is useless.

To enable a principled combination, we first transform the unbounded Compression Ratio $CR = \frac{|V_o|}{|V_n|}$ into a bounded efficiency measure, the \textbf{Vocabulary Reduction Gain (VRG)}, defined as:

\begin{equation}
VRG = 1 - \frac{1}{CR} = 1 - \frac{|V_n|}{|V_o|} = \frac{|V_o| - |V_n|}{|V_o|}
\end{equation}

VRG measures the proportion of the original vocabulary eliminated by normalization, and is bounded to $[0, 1)$: a value of 0 indicates no compression, while values approaching 1 indicate near-total vocabulary reduction. This bounded domain matches that of IRS, making the two components directly combinable.

The AES is then defined as the harmonic mean of VRG and IRS:

\begin{equation}
AES = \frac{2 \cdot IRS \cdot VRG}{IRS + VRG}
\end{equation}

The harmonic mean is chosen deliberately, as it penalizes imbalance between the two components: a high score requires both efficiency and fidelity simultaneously. Two failure modes are handled correctly by this formulation:

\begin{itemize}
    \item \textbf{Identity failure (do-nothing algorithm):} An algorithm that changes nothing achieves $CR = 1 \implies VRG = 0$, regardless of how well it preserves semantics. The AES correctly assigns a score of zero: $AES = \frac{2 \cdot IRS \cdot 0}{IRS + 0} = 0$.

    \item \textbf{Destructive failure (over-aggressive algorithm):} An algorithm that collapses the vocabulary aggressively ($VRG \to 1$) but destroys semantic content ($IRS \approx 0$) receives a near-zero AES, correctly reflecting the quality failure despite high compression.
\end{itemize}

\subsubsection{Component 5: Average Normalized Levenshtein Distance (ANLD)}
The last component is Average Normalized Levenshtein Distance (ANLD), which acts as a diagnostic measure for cases in which high-level semantic similarity (e.g. a strong IRS) hides morphological deformation. ANLD is computed by calculating the Levenshtein edit distance between each original token and its transformed version, normalizing this distance by the original token's length, and averaging the resulting values across the corpus, yielding a score that ranges from 0 (identical surface forms) to 1 (maximal transformation). This level of detail is necessary to identify the dangerous effects of over-stemming and other fine-grained distortions that macro-level metrics miss.

The Average Normalized Levenshtein Distance (ANLD) is defined as:
\begin{equation}
ANLD(V, \sigma) = \frac{1}{|V|} \sum_{w \in V} \frac{LD(w, \sigma(w))}{|w|}
\end{equation}

where:

\begin{itemize}
    \item $V$ is the vocabulary, and $|V|$ is its size,
    \item $w \in V$ is a word in the vocabulary,
    \item $|w|$ is the length of the word $w$,
    \item $\sigma(w)$ is the normalized (stemmed) form of $w$,
    \item $LD(w, \sigma(w))$ is the Levenshtein distance between $w$ and $\sigma(w)$.
\end{itemize}
A score of $0$ indicates that $w$ and $\sigma(w)$ are identical, while a score of $1$ indicates maximum transformation relative to the word length. Importantly, NormEval treats the character-level distortion gap ($\Delta$) and the downstream Model Performance Delta (MPD) as complementary but mathematically independent metrics. While MPD evaluates extrinsic, task-dependent utility (which varies wildly between classification, retrieval, or topic modeling), $\Delta$ offers a task-agnostic, intrinsic measure of morphological fidelity. This independence ensures the framework's broad generalizability across heterogeneous NLP pipelines.

\subsection{Framework API and Usage}
To facilitate reproducibility and adoption, the NormEval framework is released as an open-source Python package. Code~\ref{lst:normalization_init} illustrates the minimal interface required to execute the full evaluation suite on a user-supplied corpus.

\begin{lstlisting}[language=Python, caption={Minimal usage example of the \texttt{NormalizationEvaluator} API}, label={lst:normalization_init}]
#package installation
! pip install normeval
from normeval import NormalizationEvaluator
from sentence_transformers import SentenceTransformer
from sklearn.ensemble import RandomForestClassifier

# Define semantic and predictive models
model = SentenceTransformer('paraphrase-multilingual-MiniLM-L12-v2')
clf   = [RandomForestClassifier()]

# Initialize the evaluator
evaluator = NormalizationEvaluator(
    texts_original   = raw_texts,
    texts_normalized = normalized_texts,
    labels           = y_labels,
    classifiers      = clf,
    embedding_model  = model
)

# Run the full NormEval suite
results = evaluator.evaluate_all(lang="en")
print(results)  # CR, MPD, IRS, AES, ANLD
\end{lstlisting}

\section{Results and Discussion}
\label{sec:results}
\subsection{Multilingual Evaluation on XNLI}
To validate the generalizability of the proposed NormEval framework across typologically diverse languages, we apply it to the Cross-lingual Natural Language Inference (XNLI) benchmark \cite{conneau2018xnli}. The evaluation is restricted to the six languages for which a Snowball stemmer is available: Arabic (ar), German (de), English (en), Spanish (es), French (fr), and Russian (ru). This restriction ensures that the compression and distortion metrics reflect genuine morphological transformation rather than tokenization artifacts. A unified normalization pipeline comprising Unicode NFKC normalization, lowercasing, diacritic stripping, punctuation removal, and Snowball stemming was applied uniformly and independently to both the premise and hypothesis of each NLI pair. A multilingual sentence encoder (\texttt{paraphrase-multilingual-MiniLM-L12-v2}) was used to build NLI-aware interaction features $[\mathbf{u},\, \mathbf{v},\, |\mathbf{u}-\mathbf{v}|,\, \mathbf{u} \odot \mathbf{v}]$, and a Logistic Regression classifier trained once on the English MultiNLI training split was evaluated on each language's validation set under a zero-shot cross-lingual transfer protocol. Table~\ref{tab:multilingual_nli_results} reports the full NormEval diagnostic profile namely CR, VRG, IRS, AES, and ANLD, alongside the original and normalized macro F1-scores and the resulting MPD for each language.

\begin{table}[ht]
\centering
\caption{Multilingual NLI Evaluation on XNLI (Snowball-supported languages). $\uparrow$ = higher is better; $\downarrow$ = lower is better.}
\label{tab:multilingual_nli_results}
\small
\setlength{\tabcolsep}{4pt}
\begin{tabular}{lcccccccc}
\toprule
\textbf{Language} & \textbf{CR} $\uparrow$ & \textbf{VRG} $\uparrow$ & \textbf{IRS} $\uparrow$ & \textbf{AES} $\uparrow$ & \textbf{ANLD} $\downarrow$ & \textbf{F1\textsubscript{orig}} & \textbf{F1\textsubscript{norm}} & \textbf{MPD} \\
\midrule
Arabic (ar)  & 1.6908 & 0.4086 & 0.7787 & 0.5359 & 0.4610 & 0.6380 & 0.5711 & $-$0.0669 \\
German (de)  & 1.2810 & 0.2194 & 0.7849 & 0.3429 & 0.2976 & 0.6483 & 0.5325 & $-$0.1158 \\
English (en) & 1.3303 & 0.2483 & 0.7934 & 0.3782 & 0.2560 & 0.6767 & 0.5821 & $-$0.0945 \\
Spanish (es) & 1.6181 & 0.3820 & 0.7041 & 0.4953 & 0.3309 & 0.6624 & 0.5285 & $-$0.1339 \\
French (fr)  & 1.3053 & 0.2339 & 0.7194 & 0.3530 & 0.3334 & 0.6641 & 0.5445 & $-$0.1195 \\
Russian (ru) & 1.7473 & 0.4277 & 0.8125 & 0.5604 & 0.3285 & 0.6459 & 0.5562 & $-$0.0897 \\
\bottomrule
\end{tabular}
\end{table}
The results reveal a consistent and interpretable pattern across the six languages, which we organize into two clusters based on morphological complexity.
The following comparison is relative to each other; no constant or exact values were pre-selected. 
\textbf{High-distortion cluster (ar, es, ru):} Morphologically rich languages with productive suffixation show substantially higher compression ratios (CR $>$ 1.30) and correspondingly higher VRG values ($>$ 0.38), but also elevated ANLD scores ($>$ 0.32), signaling aggressive character-level transformation. Despite meaningful vocabulary reduction, AES values remain moderate (0.49--0.56), reflecting the penalty imposed by semantic distortion. This is accompanied by notable performance degradation, with MPD values ranging from $-$0.0669 (Arabic) to $-$0.1339 (Spanish). Critically, IRS for these languages remains moderate (0.70--0.81), suggesting that embedding-based semantic similarity alone would not flag the severity of the distortion, a finding that directly motivates the inclusion of ANLD as a complementary diagnostic in the framework. For instance, in highly inflected or root-and-pattern languages like Arabic, a high character distortion ($ANLD = 0.4610$) combined with an ostensibly stable sentence embedding ($IRS = 0.7787$) means that critical semantic tokens are being mutated. In an expert clinical decision support platform, this micro-level corruption risks stripping vital medical negation or diagnostic affixes while leaving the macro-vector unchanged.

\textbf{Intermediate cluster (en, fr, de):} English, French and German occupy an intermediate position: moderate compression (CR $\approx$ 1.28--1.33, VRG $\approx$ 0.22--0.25), moderate IRS ($\approx$ 0.72--0.79) , and lower AES values (0.34--0.38), consistent with the known behavior of Snowball stemming on Germanic languages. Meaningful performance drops (MPD of $-0.0945$, $-0.1195$, and $-0.1158$, respectively) confirm that even moderate compression carries a downstream cost.

Across all six languages, every MPD value is negative, indicating that Snowball stemming consistently reduces NLI classification performance under the zero-shot cross-lingual transfer protocol used here. Languages with higher VRG and ANLD values tend to exhibit larger performance drops, while languages with more moderate compression show smaller but still meaningful degradation. These observations are descriptive and based on a single stemming algorithm; causal claims about the relationship between normalization aggressiveness and downstream loss would require a controlled multi-algorithm comparison, which we conduct in the ablation study that follows. The XNLI results nonetheless demonstrate that the NormEval metric suite produces interpretable and consistent diagnostic profiles across six typologically distinct languages, supporting the framework's cross-lingual applicability.

\subsection{Ablation Study: Metric Necessity and Framework Validity}
The XNLI evaluation validates the framework's cross-lingual generalizability using a single normalization pipeline. However, it does not answer a more fundamental question: \emph{is each metric in the NormEval framework individually necessary, or can a subset of metrics provide equivalent diagnostic power?} To answer this, we conduct a controlled ablation study on two languages, English and Bangla, using multiple normalization algorithms per language, allowing us to isolate the contribution of each metric component.

\subsubsection{Ablation Design}
The ablation study follows a \emph{metric-removal} design: we systematically evaluate what information is lost when each metric is excluded from the framework. Specifically, we define five ablation variants of the full NormEval framework (denoted \textbf{Full}):

\begin{itemize}
    \item \textbf{w/o CR}: Framework without Compression Ratio — cannot assess macro-level vocabulary efficiency.
    \item \textbf{w/o MPD}: Framework without Model Performance Delta — cannot assess downstream task impact.
    \item \textbf{w/o IRS}: Framework without Information Retention Score — cannot assess semantic preservation.
    \item \textbf{w/o AES}: Framework without Algorithm Effectiveness Score — loses the integrated efficiency–preservation balance metric.
    \item \textbf{w/o ANLD}: Framework without Average Normalized Levenshtein Distance — loses the micro-level morphological safety gate.
\end{itemize}
For each variant, we ask: \emph{would a practitioner using only the remaining metrics reach the same conclusion about algorithm quality?} The study is conducted on the sentence complexity corpus (English: 4 algorithms; Bangla: 4 algorithms) and the sentiment corpus (3 algorithms), providing 11 algorithm-dataset combinations as test cases.

\subsubsection{Ablation Result 1: Necessity of CR}
Table~\ref{tab:comparison_multirow_final} reports vocabulary sizes and TF-IDF feature dimensionalities before and after normalization for all algorithms.

\begin{table}[ht]
\centering
\caption{Compression Ratio (CR) across algorithms and languages. Removing CR from the framework leaves practitioners unable to distinguish aggressive from conservative normalizers at the macro level.}
\label{tab:comparison_multirow_final}
\begin{tabular}{llrr}
\toprule
\textbf{Language} & \textbf{Method} & \textbf{VocabSize (CR)} & \textbf{TF-IDF Features (CR)} \\
\midrule
\multirow{5}{*}{English} 
 & No processing  & 2175 (1.00)   & 1756 (1.00) \\
 & Porter         & 1563 (1.39)   & 1398 (1.26) \\
 & Snowball       & 1555 (1.40)   & 1385 (1.27) \\
 & Wordnet        & 1836 (1.18)   & 1623 (1.08) \\
 & SpaCy          & 1556 (1.40)   & 1358 (1.29) \\
\midrule
\multirow{5}{*}{Bangla} 
 & No processing & 2956 (1.00)   & 682 (1.00) \\
 & BNLTK         & 2227 (1.33)   & 603 (1.13) \\
 & BaNeL         & 2542 (1.16)   & 664 (1.03) \\
 & BanLemma      & 2557 (1.16)   & 652 (1.05) \\
 & BanglaLem     & 2459 (1.20)   & 638 (1.07) \\
\bottomrule
\end{tabular}
\end{table}

In English, Snowball and SpaCy achieve the strongest compression (CR = 1.40; $\sim$29\% vocabulary reduction; 27–29\% TF-IDF reduction), while WordNet lemmatization provides only moderate compression (CR = 1.18), consistent with its conservative semantics-preserving design. For Bangla, BNLTK achieves the highest vocabulary reduction (CR = 1.33), while BaNeL and BanLemma show lower reduction (CR = 1.16). \textbf{Ablation finding:} Without CR, all four English stemmers would appear equivalent in downstream performance (see Section~\ref{sec:abl_mpd}), making it impossible to distinguish computationally efficient algorithms from conservative ones. CR is therefore a necessary component for resource-constrained deployment decisions.

\subsubsection{Ablation Result 2: Necessity of MPD}
\label{sec:abl_mpd}

\begin{figure*}[ht]
    \centering
    \includegraphics[width=0.9\linewidth]{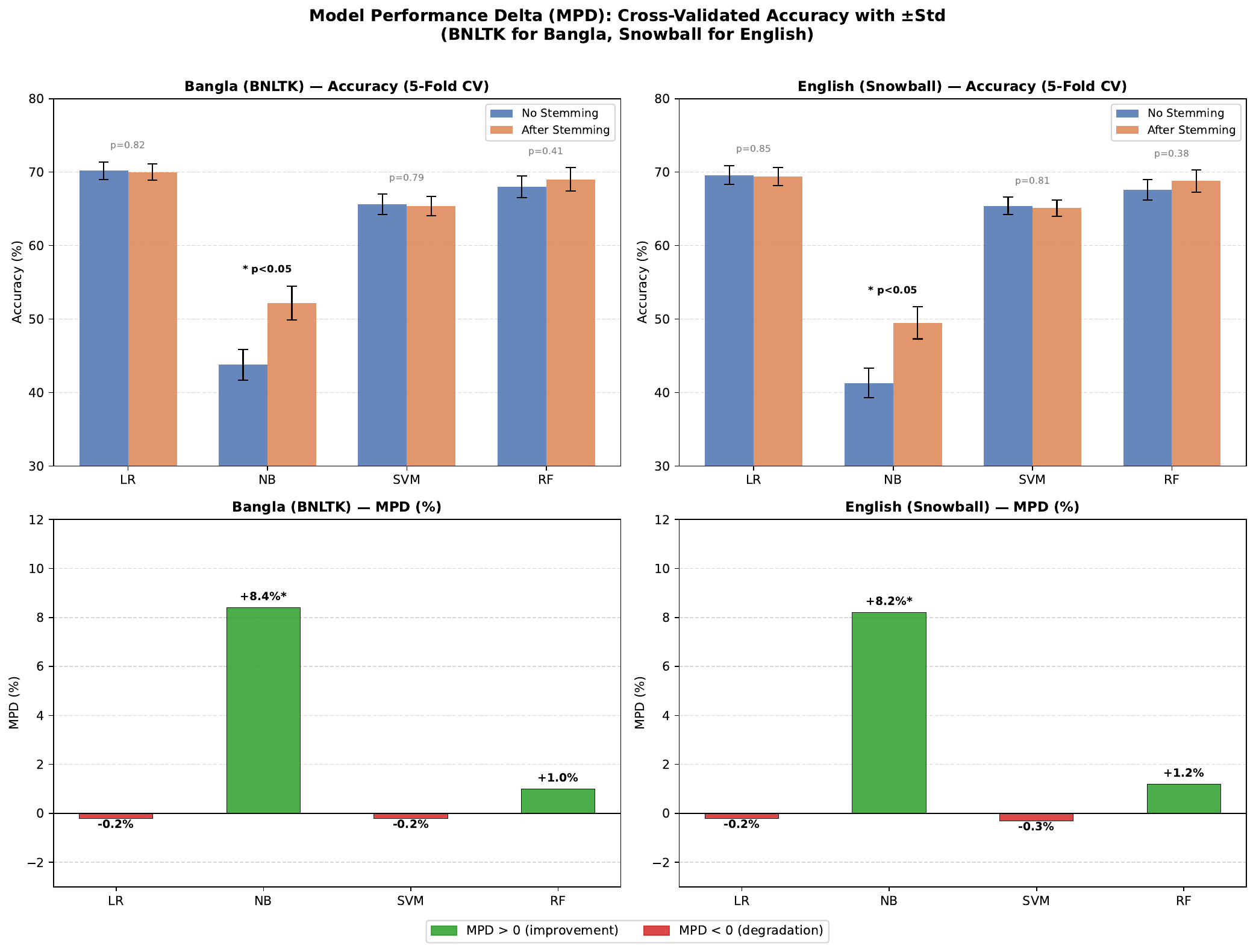}
    \caption{Model Performance Delta (MPD) across classifiers for Snowball (English) and BNLTK (Bangla).}
    \label{fig:mpd}
\end{figure*}

Figure~\ref{fig:mpd} reports the Model Performance Delta for each classifier across Snowball (English) and BNLTK (Bangla).

Across both languages, all MPD values are statistically insignificant (p $>$ 0.05) except Naive Bayes, with Naive Bayes showing the largest post-stemming improvement due to its dependence on vocabulary sparsity. \textbf{Ablation finding:} This result reveals that MPD, when used in isolation, is an unreliable performance metric, a conclusion that would be missed without its inclusion in the framework. The key contribution of this ablation is redefining MPD's role: rather than a \emph{performance metric}, it should function as a \emph{hygiene metric}, a pass/fail gate to prevent performance degradation. Without MPD, a practitioner cannot verify that normalization does not harm downstream tasks, even when CR and IRS appear favorable.

\subsubsection{Ablation Result 3: Necessity of IRS and AES}

Table~\ref{tab:SESIRS_modified} reports VRG, IRS, and AES for the two most aggressive algorithms identified in Section~\ref{sec:abl_mpd}.

\begin{table}[ht]
\centering
\caption{VRG, IRS, and AES for the highest-compression algorithms. Removing IRS and AES from the framework leaves CR as the sole efficiency signal, which overestimates algorithm quality.}
\label{tab:SESIRS_modified}
\begin{tabular}{@{}lcccc@{}}
\toprule
\textbf{Algorithm} & \textbf{CR} & \textbf{VRG} & \textbf{IRS} & \textbf{AES} \\
\midrule
BNLTK (Bangla)      & 1.33 & 0.2481 & 0.87 & 0.3861 \\
Snowball (English)  & 1.40 & 0.2857 & 0.73 & 0.4107 \\
\bottomrule
\end{tabular}
\end{table}

Despite BNLTK achieving higher semantic retention (IRS = 0.87 vs. 0.73), its lower VRG (0.2481 vs. 0.2857) results in a lower AES (0.3861 vs. 0.4107), meaning AES correctly penalizes BNLTK for its comparatively weaker compression efficiency. However, both algorithms receive moderate AES scores that do not expose the severity of BNLTK's character-level destruction. \textbf{Ablation finding:} A framework relying only on \{CR, MPD, IRS, AES\} would fail to detect fine-grained morphological distortion, since all four metrics are either vocabulary-level or embedding-based and are insensitive to character-level transformation. This finding directly motivates the necessity of ANLD, as demonstrated next.

\subsubsection{Ablation Result 4: Necessity of ANLD,  The Safety Gate Hypothesis}

\begin{figure*}[ht]
    \centering
    \includegraphics[width=0.9\linewidth]{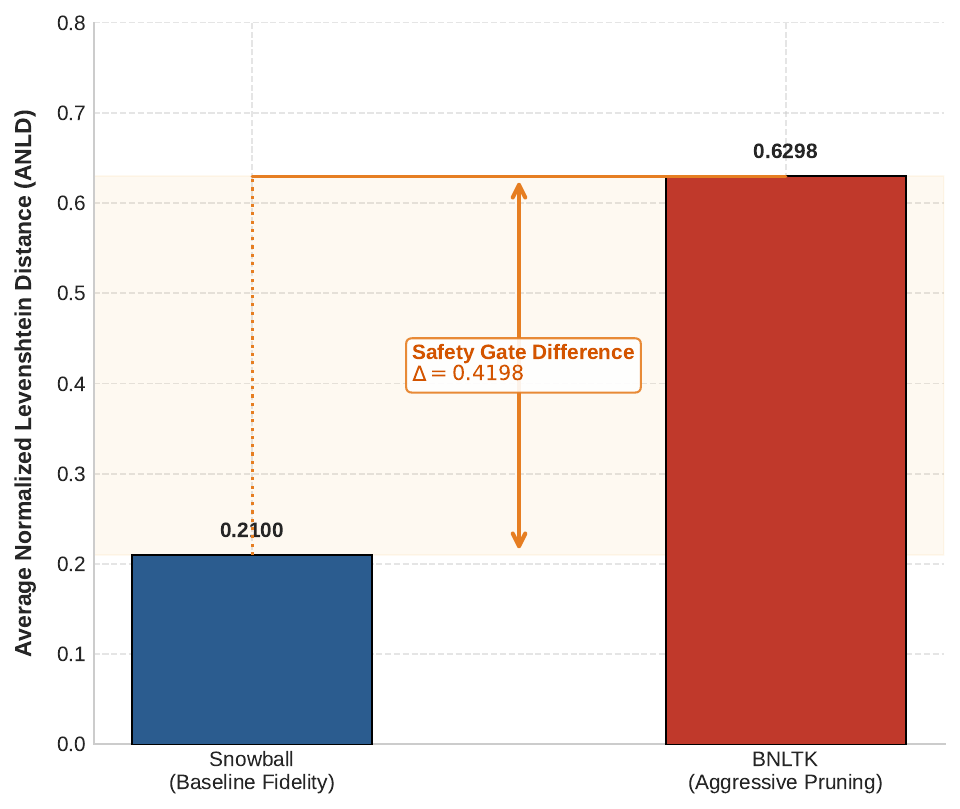}
    \caption{ANLD values and over-stemming error examples. BNLTK's high ANLD (0.63) exposes destructive morphological truncation that IRS (0.87) fails to detect.}
    \label{fig:anld_errors}
\end{figure*}

Figure~\ref{fig:anld_errors} reports ANLD values alongside over-stemming examples for both algorithms.

\begin{figure*}[ht]
    \centering
    \includegraphics[width=0.9\linewidth]{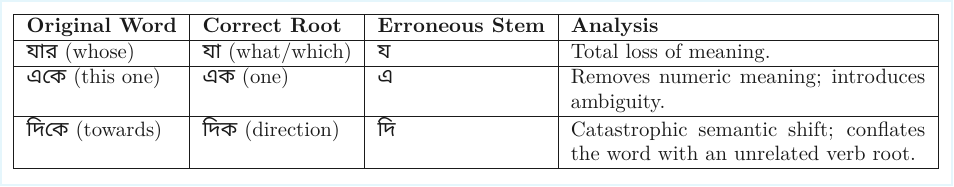}
    \caption{Examples of over-stemming errors in Bangla captured by ANLD but missed by IRS.}
    \label{fig:overstem_example}
\end{figure*}

The ANLD for Snowball is 0.21, while the ANLD for BNLTK is 0.63. Despite BNLTK's higher IRS, its ANLD reveals that it frequently reduces Bangla words to fragments that do not correspond to valid linguistic units, a form of destructive over-stemming that embedding-based metrics cannot detect because contextual embeddings smooth over surface-form distortions. Figure~\ref{fig:overstem_example} provides concrete examples of this phenomenon.

Here, the ANLD threshold $\tau$ or baseline is not a universal constant but a task-dependent reference derived from our evaluation set.

\textbf{Ablation finding:} Removing ANLD from the framework would cause a practitioner to prefer BNLTK over Snowball based on IRS and AES alone, the opposite of the correct conclusion. This confirms the \textbf{Safety Gate Hypothesis}: ANLD is a necessary and non-redundant component that detects morphological distortion invisible to macro-level metrics.

\subsubsection{Ablation Summary}

Table~\ref{tab:ablation_summary} consolidates the ablation findings across all metric-removal variants, showing the diagnostic failure that results from each omission.

\begin{table}[ht]
\centering
\caption{Ablation Summary: Diagnostic failure introduced by removing each NormEval component. \checkmark = correct conclusion reached; \ding{55} = incorrect or incomplete conclusion.}
\label{tab:ablation_summary}
\begin{tabular}{lccccc}
\toprule
\textbf{Framework Variant} & \textbf{Efficiency} & \textbf{Downstream} & \textbf{Semantic} & \textbf{Morphological} & \textbf{Overall} \\
                           & \textbf{Assessment} & \textbf{Safety}     & \textbf{Fidelity} & \textbf{Safety}        & \textbf{Ranking} \\
\midrule
Full NormEval              & \checkmark & \checkmark & \checkmark & \checkmark & \checkmark \\
w/o CR                     & \ding{55}  & \checkmark & \checkmark & \checkmark & \ding{55}  \\
w/o MPD                    & \checkmark & \ding{55}  & \checkmark & \checkmark & \ding{55}  \\
w/o IRS                    & \checkmark & \checkmark & \ding{55}  & \checkmark & \ding{55}  \\
w/o AES                    & \checkmark & \checkmark & \ding{55}  & \checkmark & \ding{55}  \\
w/o ANLD                   & \checkmark & \checkmark & \checkmark & \ding{55}  & \ding{55}  \\
\midrule
CR only (prior work)       & \checkmark & \ding{55}  & \ding{55}  & \ding{55}  & \ding{55}  \\
Accuracy only (prior work) & \ding{55}  & \checkmark & \ding{55}  & \ding{55}  & \ding{55}  \\
\bottomrule
\end{tabular}
\end{table}

The ablation summary demonstrates that every component of NormEval is individually necessary: removing any single metric causes at least one dimension of the evaluation to fail, leading to incorrect algorithm rankings. The last two rows contextualize the improvement over prior work: single-metric approaches used in the literature (CR-only or accuracy-only) fail on four out of five diagnostic dimensions, whereas the full NormEval framework succeeds on all five.

\subsection{Downstream Task Robustness and Safety Gate Calibration}
To further validate the framework under domain shift and class imbalance, we evaluate 
on the Bangla sentiment corpus \cite{Biswas2025} (9,163 examples; Positive: 4,139; 
Negative: 2,826; Neutral: 2,189). This corpus introduces greater variation in content 
and language style than the complexity corpus, providing a more realistic, uncontrolled 
setting for testing the robustness of the proposed metrics.

\begin{table}[ht]
\caption{Full NormEval metrics on the Bangla sentiment corpus.}
\centering
\begin{tabular}{|>{\raggedright\arraybackslash}p{2.5cm}|c|c|c|c|l|c|c|c|}
\hline
\multirow{2}{*}{Algorithm}
& Compression
& VRG
& Information Retention
& \multirow{2}{*}{AES}
& \multicolumn{3}{c|}{Model F1}
& \multirow{2}{*}{ANLD} \\
\cline{6-8}
& Ratio (CR)
&
& Score (IRS)
&
& Model
& Org
& Norm
& \\
\hline
\multirow{4}{*}{\parbox{3.2cm}{\raggedright
Snowball\\\cite{porter2001snowball}}}
& \multirow{4}{*}{1.64}
& \multirow{4}{*}{0.39}
& \multirow{4}{*}{0.80}
& \multirow{4}{*}{0.52}
& RF  & 68.21 & 69.59 & \multirow{4}{*}{0.14} \\
& & & & & LR  & 66.10 & 64.14 & \\
& & & & & MNB & 65.33 & 65.81 & \\
& & & & & SVM & 65.17 & 65.76 & \\
\hline
\multirow{4}{*}{\parbox{3.2cm}{\raggedright
BNLTK\\\cite{bnltk}}}
& \multirow{4}{*}{1.90}
& \multirow{4}{*}{0.47}
& \multirow{4}{*}{0.88}
& \multirow{4}{*}{0.62}
& RF  & 66.42 & 64.61 & \multirow{4}{*}{0.26} \\
& & & & & LR  & 59.21 & 58.87 & \\
& & & & & MNB & 59.76 & 58.21 & \\
& & & & & SVM & 58.52 & 57.22 & \\
\hline
\multirow{4}{*}{\parbox{3.2cm}{\raggedright
BanLemma\\\cite{sadia2019n}}}
& \multirow{4}{*}{1.61}
& \multirow{4}{*}{0.38}
& \multirow{4}{*}{0.91}
& \multirow{4}{*}{0.54}
& RF  & 65.54 & 67.10 & \multirow{4}{*}{0.18} \\
& & & & & LR  & 60.21 & 62.33 & \\
& & & & & MNB & 53.72 & 57.11 & \\
& & & & & SVM & 58.41 & 65.72 & \\
\hline
\label{tab:final_comparison}
\end{tabular}
\end{table}

The results in Table~\ref{tab:final_comparison} reveal a critical dissociation between AES and morphological safety, which the full NormEval framework is uniquely positioned to expose. BNLTK achieves the highest AES (0.62), driven by its aggressive compression (CR = 1.90, VRG = 0.47) combined with strong semantic retention (IRS = 0.88). However, its ANLD (0.26) exceeds the safety threshold, and this morphological destruction manifests directly in downstream performance: BNLTK produces negative MPD across all four classifiers (mean MPD = $-$1.25\%), confirming that AES alone is an insufficient guide for algorithm selection. Snowball presents a near-neutral profile (mean MPD = $+$0.12\%), with modest gains 
for RF, MNB, and SVM offset by a drop in LR ($-$1.96\%), consistent with its 
moderate compression (CR = 1.64) and safe ANLD (0.14). BanLemma provides the 
strongest evidence for the framework's practical utility: despite a lower AES (0.54) than BNLTK, its conservative compression (CR = 1.61), highest semantic retention (IRS = 0.91), and safe ANLD (0.18) yield consistent downstream gains across all classifiers (mean MPD = $+$3.59\%), including a 7.31 percentage point improvement in SVM macro F1-score (58.41\% $\to$ 65.72\%). This is the clearest demonstration of the Safety Gate Hypothesis: the algorithm with the highest AES is not the best choice, and this conclusion is only reachable when ANLD is included in the framework.\\
\textbf{Safety Gate Threshold Calibration.} The ANLD threshold $\tau$ is not a 
universal constant but a task-dependent reference derived from the practitioner's 
evaluation set. In this case, we can use two calibration strategies:
\begin{itemize}
    \item \textbf{Compression-first calibration:} $\tau$ is set to the ANLD of the 
    algorithm with the highest CR in the comparison set, the most aggressive 
    compressor. Any algorithm equal to or exceeding this value is flagged as disproportionately 
    destructive relative to its compression gain. In our experiments, BNLTK 
    (CR = 1.90, ANLD = 0.26) defines this upper bound (inclusive).

    \item \textbf{Semantic-retention calibration:} $\tau$ is set to the ANLD of the 
    algorithm with the highest IRS, the most semantically conservative option. 
    This is the recommended baseline for nuance-sensitive tasks such as sentiment 
    analysis or question answering. In our experiments, BanLemma (IRS = 0.91, 
    ANLD = 0.18) serves as this reference; any algorithm with ANLD $>$ 0.18 is 
    flagged as morphologically unsafe relative to the best semantically conservative 
    alternative.
\end{itemize}
Crucially, both strategies converge on the same conclusion in our data: BNLTK 
(ANLD = 0.26) is flagged under both calibrations, while Snowball (ANLD = 0.14) 
and BanLemma (ANLD = 0.18) are within safe limits. This convergence strengthens 
the Safety Gate Hypothesis: the threshold is robust to calibration strategy choice 
when the algorithm comparison set is sufficiently diverse. Practitioners working in 
new language settings should re-derive $\tau$ from their own algorithm comparison 
set before applying the safety gate. In addition to this, we do not attempt to mathematically bind the structural distortion gap ($\Delta$ ANLD) to the Model Performance Delta (MPD). Because MPD is highly task-dependent, varying widely depending on whether the downstream application is classification, topic modeling, or retrieval, a universal mathematical mapping is not feasible. Instead, we treat $\Delta$ as a purely intrinsic, comparative gauge. It allows us to quantify exactly how much more aggressively one algorithm mutates word formations compared to another, functioning as a structural ``hygiene check'' that is independent of the chosen downstream task.

\section{Conclusion}
\label{sec:conclusion}
Text normalization is a foundational step in NLP pipelines, yet its evaluation has remained methodologically fragmented; most existing work relies on a single metric, either vocabulary compression or downstream performance, neither of which is sufficient to characterize the full impact of normalization on linguistic quality. This paper addresses that gap by introducing NormEval, a unified, language-agnostic evaluation framework that simultaneously assesses normalization methods across three complementary dimensions: macro-level computational efficiency, downstream task utility, and micro-level morphological fidelity.

The framework operationalizes five metrics: Compression Ratio (CR), Model Performance Delta (MPD), Information Retention Score (IRS), Algorithm Effectiveness Score (AES), and Average Normalized Levenshtein Distance (ANLD), each targeting a distinct and non-redundant aspect of normalization quality. The controlled ablation study on English and Bangla corpora demonstrates that removing any single metric from the framework causes at least one dimension of the evaluation to fail, leading to incorrect algorithm rankings. Most critically, the study establishes the \textbf{Safety Gate Hypothesis}: embedding-based metrics such as IRS can assign high semantic similarity scores to texts that have undergone destructive morphological truncation, and only ANLD reliably exposes this class of failure. This finding has direct practical implications; a practitioner relying solely on IRS and AES would incorrectly prefer BNLTK over BanLemma for Bangla, whereas the full NormEval framework correctly identifies BanLemma as the superior algorithm based on its balanced profile (CR = 1.61, IRS = 0.91, AES = 0.54, ANLD = 0.18).

The multilingual evaluation on the XNLI benchmark further validates the framework's cross-lingual generalizability across six typologically diverse languages. The results reveal a consistent and interpretable pattern: morphologically rich languages such as Arabic, Spanish, and Russian exhibit high compression but also elevated ANLD and substantial downstream performance degradation. This cross-lingual landscape confirms that the relationship between normalization aggressiveness and linguistic safety is language-specific and cannot be captured by any single metric.

Beyond diagnosis, NormEval provides a practical decision-support tool. The framework supports iterative ANLD threshold calibration, allowing practitioners to dynamically adjust the safety gate based on task requirements: a relaxed threshold for information retrieval tasks where efficiency is paramount, and a stricter threshold for nuance-sensitive applications such as sentiment analysis, where morphological integrity directly affects predictive performance. The downstream robustness analysis on the Bangla sentiment corpus confirms this principle empirically, with BanLemma yielding a 7.31 percentage point SVM macro F1-score gain precisely because its ANLD remained within safe limits.

Taken together, these findings establish NormEval as a standardized, reproducible methodology for the NLP community to evaluate, compare, and select normalization algorithms in a principled and linguistically informed manner. NormEval lays the groundwork for more reliable and linguistically informed intelligent systems, from clinical decision support and legal document analysis to sentiment-aware recommendation engines, where the silent corruption of meaning by poorly evaluated normalization carries real operational cost.

\section{Limitations and Future Work}
\label{sec:limitations}
Despite the contributions outlined above, several limitations of the current study warrant acknowledgment and motivate future research directions.

\textbf{Scope of normalization methods.} The current study focuses on stemming and lemmatization as the primary normalization strategies. Other normalization techniques, such as synonym replacement, spelling correction, abbreviation expansion, and subword tokenization, are not evaluated within the NormEval framework. While the five metrics are theoretically applicable to any token-level transformation, their behavior under these alternative normalization paradigms has not been empirically validated, and the ANLD safety gate thresholds established in this study may not transfer directly.

\textbf{Transformer-based IRS limitations.} The Information Retention Score relies on transformer-based sentence embeddings, which are known to be insensitive to fine-grained morphological variation, a limitation that is precisely why ANLD was introduced as a complementary metric. However, the choice of embedding model (BanglaBERT for Bangla, DistilBERT for English) introduces model-specific biases, and IRS values may vary across embedding architectures. Future work should investigate the sensitivity of IRS to embedding model choice and explore character-level or morpheme-aware embeddings as alternatives.

\textbf{Language coverage.} The controlled ablation study is limited to separate datasets on English and Bangla. The ANLD safety gate thresholds and the specific failure modes identified (e.g., BNLTK's destructive truncation) are language-specific findings. Extending the ablation study to morphologically rich languages such as Turkish, Finnish, or Arabic, where agglutinative morphology poses even greater normalization challenges, would strengthen the generalizability of the framework's diagnostic conclusions.

\textbf{Dataset domain and size.} The primary evaluation corpus (sentence complexity classification) contains approximately 2,659 examples, and the auxiliary sentiment corpus contains 9,163 examples. While these sizes are sufficient for the controlled experimental design of this study, they may not fully represent the distributional diversity of real-world NLP corpora. Performance of normalization algorithms, and the reliability of the proposed metrics, may differ on larger, noisier, or domain-specific datasets such as clinical text, legal documents, or social media data.

\textbf{ANLD threshold generalizability.} The ANLD threshold should be treated as a language- and task-specific starting point rather than a universal constant. A systematic study of ANLD threshold calibration across multiple languages, domains, and downstream tasks is needed to establish more robust guidelines for practitioners.

\textbf{Statistical validation scope.} Statistical significance testing (p-values) is applied to MPD in this study. Future work should extend formal statistical validation, including effect size measures and confidence intervals, to the other metrics, particularly IRS and ANLD, to provide stronger guarantees about the reliability of metric differences across algorithms.

\textbf{PyPI Package Limitations.} The current NormEval package relies on whitespace-based tokenization and therefore does not natively support languages that use character-level or script-continuous writing systems, such as Chinese, Thai, or Japanese. For such languages, a dedicated segmentation tool (e.g., \texttt{jieba} for Chinese, \texttt{PyThaiNLP} for Thai) must be applied as a preprocessing step before passing text to the evaluator. Extending the package to incorporate language-aware tokenization backends and validating the full metric suite on character-segmented and agglutinative languages is a priority for future development.

\textbf{MPD task coverage.} The Model Performance Delta is defined generically as the difference in any task-appropriate performance measure before and after normalization. However, the empirical instantiation in this study is limited to text classification (macro-F1 on sentence complexity and sentiment corpora). The behavior of MPD under other downstream task types, including named entity recognition, machine translation (BLEU), abstractive summarization (ROUGE), and question answering (F1/EM), has not been evaluated. It is plausible that normalization affects these tasks differently: for instance, aggressive stemming may harm NER by destroying entity surface forms while having a negligible impact on document-level sentiment classification. Extending the MPD case study to a broader range of NLP tasks is an important direction for future work.

\textbf{Safety Gate threshold generalizability.} The ANLD threshold $\tau$ is 
currently derived from either the most aggressively compressing or the most semantically conservative algorithm in the comparison set, making it corpus- and language-dependent. Its behavior on morphologically complex languages beyond Bangla, Arabic, and Russian has not been validated, and a universal threshold calibration procedure remains an open research question.

Future work will focus on three directions: (1) extending NormEval to cover a broader range of normalization techniques, including neural and subword-based methods; (2) developing language-adaptive ANLD thresholds through large-scale cross-lingual benchmarking; and (3) releasing NormEval as an open-source evaluation toolkit with standardized interfaces for integration into existing NLP pipelines, enabling reproducible and comparable normalization evaluation across the research community.

\section{Funding Information}
N/A

\section{Data Availability}
All datasets, experimental code are available \url{https://anonymous.4open.science/r/Stemming-A953/}. 
Reproducible code capsules used in this study are hosted publicly and will be made accessible via a live link immediately upon acceptance of the manuscript. The URL has been omitted in this version to preserve the double-blind review process.

% and reproducible code capsules used in this study are publicly available at: \url{https://colab.research.google.com/drive/1CU-JS9LhdwEyZoH0GXsxL3iFkVlrz5P9?usp=sharing} 

\section{Software Availability}
Full documentation, usage examples, and the reproducible code capsule are available at:
\url{https://anonymous.4open.science/r/normeval-C74F}
% The NormEval framework is released as an open-source Python package and can be installed via the Python Package Index (PyPI): \texttt{pip install normeval}

\section{CRediT Author Statement}

Md Abdullah Al Kafi: Conceptualization, Methodology, Writing – original draft\\
Raka Moni: Writing - Original Draft, Writing - Review \& Editing, Validation\\
Walayat Hussain: Supervision.

\section{Declaration of Generative AI and AI-assisted technologies in the writing process}
During the preparation of this work, the author(s) used Gemini (Google) to refine the clarity of the English text. After using this tool/service, the author(s) reviewed and edited the content as needed and take(s) full responsibility for the content of the publication.

\bibliographystyle{cas-model2-names}
\bibliography{cas-refs}
\end{document}